# DexDeepFM: Ensemble Diversity Enhanced Extreme Deep Factorization Machine Model

LING CHEN*

Zhejiang University

HONGYU SHI

Zhejiang University

Predicting user positive response (e.g., purchases and clicks) probability is a critical task in Web applications. To identify predictive features from raw data, the state-of-the-art extreme deep factorization machine model (xDeepFM) introduces a new interaction network to leverage feature interactions at the vector-wise level explicitly. However, since each hidden layer in the interaction network is a collection of feature maps, it can be viewed essentially as an ensemble of different feature maps. In this case, only using a single objective to minimize the prediction loss may lead to overfitting and generate correlated errors. In this paper, an ensemble diversity enhanced extreme deep factorization machine model (DexDeepFM) is proposed, which designs the ensemble diversity measure in each hidden layer and considers both ensemble diversity and prediction accuracy in the objective function. In addition, the attention mechanism is introduced to discriminate the importance of ensemble diversity measures with different feature interaction orders. Extensive experiments on three public real-world datasets are conducted to show the effectiveness of the proposed model.

## 1 INTRODUCTION

Predicting the positive response (e.g., purchases and clicks) probability of a user on a specific item based on particular contextual information is a critical task in Web applications, including recommendation systems [1-4] and online advertising [5]. The key challenge in prediction is how to identify effective features from the abundant raw data. To take a toy example, since the preferences of a little boy and a middle-aged woman may be greatly different, considering the interaction between feature "gender" and feature "age group" may be of great help in prediction. Therefore, feature interactions are widely leveraged to predict the positive response probability of users, which have been proved effective in various applications, e.g., Location recommendation [6], App recommendation [7-8], and click-through rate prediction [9].

* This work was funded by the National Key Research and Development Program of China (No. 2018YFB0505000) and the Fundamental Research Funds for the Central Universities (No. 2020QNA5017).
Authors' addresses: L. Chen (Corresponding author), College of Computer Science and Technology, Alibaba-Zhejiang University Joint Research Institute of Frontier Technologies, Zhejiang University, Hangzhou 310027, China; email: lingchen@cs.zju.edu.cn; H. Shi, College of Computer Science and Technology, Zhejiang University, Hangzhou 310027, China; email: shihongyu@cs.zju.edu.cn.

Existing efforts on feature interactions can be classified into two categories, traditional models [10-12] and deep learning based models [8-9, 14-15]. The former category mainly includes feature engineering and factorization based models. Feature engineering based models [10] usually focus on manually designing feature interactions according to specific domain knowledge, i.e., creating a new feature by combining two or more features. However, it is hard to identify all potential feature interactions manually when the feature space is large. Factorization based models, e.g., factorization machine (FM) [11] and field-aware factorization machine (FFM) [12], exploit the idea of matrix factorization to reduce model parameters largely when conducting feature interactions, which model feature interactions as the inner products of latent vectors by projecting features into a low dimension latent space. Generally, due to the high computational cost, only second-order feature interactions are exploited, which may limit the representation power of these models. Although researchers try to extend them to arbitrary higher-order feature interactions [13], it still relies on a large number of parameters, and may bring noises with ineffective interactions [16].

Since deep learning has shown its great capability in various areas, e.g., computer vision, speech recognition, and natural language processing in recent years, some researchers have tried to apply deep learning to implement feature interactions, which can be further divided into two categories, the single structure based models [14-15] and hybrid structures based models [8-9]. In the former, with a pre-trained FM layer or a product layer, deep neural networks (DNNs) are directly applied to learn sophisticated feature interactions automatically, e.g., factorization-machine supported neural networks (FNN) [14] and product-based neural networks (PNN) [15]. However, these models pay more attention to high-order interactions, and ignore low-order feature interactions that are also important for predicting the positive response probability of users. In the latter, hybrid structures, which combine a shallow part (e.g., feature engineering or FM) and a deep part, are applied to capture both low-order and high-order feature interactions, e.g., Wide & Deep [8] and DeepFM [9]. The problem is that all these models rely on DNNs to learn high-order feature interactions at the dimension-wise level, which are in an implicit manner and poorly interpretable.

To alleviate these issues, Lian et al. [17] proposed the extreme deep factorization machine model (xDeepFM), which replaces the shallow part with a new interaction network to learn feature interactions at the vector-wise level. In addition, the order of feature interactions increases with the depth of the interaction network in an explicit manner. Specifically, the embedding vector is reshaped to an initial matrix for vector-wise feature interactions. The output of the next hidden layer, which is a collection of different feature maps, is dependent on the initial matrix and the output of the last hidden layer. Then, the sum pooling is applied on each feature map of the hidden layer, and all the sum pooling results from hidden layers are concatenated before being sent to the output unit. Finally, the objective function is to minimize the prediction loss of the task. Although xDeepFM has made great achievements over the state-of-the-art models, there still exist some deficiencies to be improved. Since the output of each hidden layer in the interaction network is a collection of different feature maps, it can be viewed essentially as an ensemble of different feature maps. However, only optimizing the prediction accuracy in the objective function of the task and overlooking the diversity across different feature maps would inevitably lead to overfitting and not generalize well, i.e., generating correlated errors, which means that the error on the unseen data would be large [18].

In recent years, ensemble learning, which contains a set of base learners, has been proven theoretically and empirically to outperform a single learner in various areas [19-22]. Meanwhile, ensemble diversity measures have shown its effectiveness to ensure that all the base learners craft uncorrelated errors [23].

Specifically, the key to build a good ensemble is to construct base learners that perform better than random guessing individually (i.e., accuracy requirement) and make errors on different parts of the training data (i.e., diversity requirement) [24]. In other words, it should have an accurate and diverse set of base learners, and the trade-off between accuracy and diversity becomes important.

In order to address the aforementioned problems and exploit the merits of ensemble learning, we propose an ensemble Diversity Enhanced eXtreme Deep Factorization Machine model (DexDeepFM), which is based on the work of xDeepFM [17]. Inspired by the idea of ensemble learning, we regard the different feature maps of the same hidden layer in the interaction network as the outputs of different base learners by analogy, which are expected to be as diverse as possible to achieve good ensembles. Specifically, the ensemble diversity measure is introduced in each hidden layer, and ensemble diversity and prediction accuracy are jointly considered in the objective function. In addition, the attention mechanism is employed to learn the attentive weights for the ensemble diversity measures of each hidden layer. The main contributions of this paper are summarized as follows:

1) We propose DexDeepFM, an ensemble diversity enhanced extreme deep factorization machine model, which exploits the merits of ensemble learning and considers both ensemble diversity and prediction accuracy in the objective function.

2) We design the ensemble diversity measure in each hidden layer, which ensures that distances between different feature maps are as far as possible. In addition, the attention mechanism is employed to discriminate the contributions of ensemble diversity measures with different feature interaction orders.

3) We conduct extensive experiments on three public real-world datasets, and experimental results show the effectiveness of the proposed model.

The rest of this paper is organized as follows: Section 2 provides a review of the related work. Section 3 introduces the proposed model. The experimental settings and results are given in Section 4. Finally, we present the concluding remarks and future work in Section 5.

## 2 RELATED WORK

In this section, a review of the related work is presented, which includes traditional feature interaction models, deep learning based feature interaction models, ensemble learning, and attention mechanism.

### 2.1 Traditional Feature Interaction Models

Traditional feature interaction models mainly include feature engineering and factorization based models. On the one hand, feature engineering based models [25-26] focus on manually crafting a new feature by combining two or more related features according to specific domain knowledge. Lian et al. [25] elaborately crafted features by interacting users' profile with jobs' profile, e.g., a user's industry with a job's industry, to indicate the relevance between users and jobs in recommendation. Jahrer et al. [26] crafted interactive features for online searching advertisements, e.g., gender-position (i.e., the position of advertisements in a query session) and age-depth (i.e., the number of advertisements displayed to users in a query session). However, these studies come at a high cost when the feature space is large, because it is not a trivial thing to identify all useful feature interactions.

On the other hand, researchers start to exploit factorization based models [11-12] to address aforementioned problems. Based on the idea of matrix factorization, Rendle [11] proposed a FM model, which projects each feature into a low dimension latent vector, and learns feature interactions from the inner products of latent vectors. Based on FM, Juan et al. [12] further presented a FFM model, which learns different vectors with regard to specific fields for each feature. In addition, Xiao et al. [16] proposed an attentional FM model, which introduces the attention mechanism and models the importance of all feature interactions. However, these studies only exploit low-order feature interactions, which may limit their representation power. Although Blondel et al. [13] extended FM to arbitrary high-order feature interactions by an efficient algorithm, it may introduce noises and reduce the overall performance [16], because all effective and ineffective feature interactions are considered in the model. In addition, it still relies on a large number of parameters.

## 2.2 Deep Learning Based Feature Interaction Models

In recent years, deep learning has been successfully applied in many areas, e.g., computer vision, speech recognition, and natural language processing. Since DNNs can automatically learn discriminative feature representations from training data, researchers have also tried to develop deep learning based feature interaction models [8-9, 14-15]. Zhang et al. [14] proposed a FNN model, which is based on a pre-trained FM and applies DNNs to learn high-order feature interactions. Similarly, He et al. [27] introduced a Bi-Interaction pooling operation layer to deepen FM under the neural network framework. Qu et al. [15] built a PNN model, which incorporates a product layer before the fully connected layer to capture interactive patterns between features. However, these studies pay more attention to learning sophisticated high-order feature interactions, and ignore the importance of low-order ones.

As an alternative solution, models [8-9, 17] with hybrid structures combining shallow and deep parts are proposed to capture both low-order and high-order feature interactions. Cheng et al. [8] proposed a Wide & Deep model, which combines the benefits of memorization and generalization for recommendation by jointly training the wide linear model and DNNs. Instead of using feature engineering, Guo et al. [9] further proposed a DeepFM model, which replaces the linear model in Wide & Deep [8] with the architecture of FM. In addition, Yu et al. [28] proposed an input-aware factorization machine model (IFM), which learns a unique input-aware factor for the same feature in different instances via a neural network and yields a competitive performance. However, there are inherent limitations in these studies. DNNs learn feature interactions at the dimension-wise level, which is hard to be interpreted explicitly. To this end, Lian et al. [17] proposed an xDeepFM model, which designs a new interaction network to replace the shallow part in DeepFM [9] and learns explicit feature interactions at the vector-wise level, which is detailed in Section 1.

The main difference between xDeepFM and the proposed model DexDeepFM is that an ensemble diversity enhanced interaction network is introduced to replace the original one. Similarly, the output of the next hidden layer is generated from the initial matrix and the output of the last hidden layer. However, we regard the different feature maps in the hidden layer as the outputs of different base learners by analogy, and the ensemble diversity measure is designed to ensure that distances between different feature maps are as far as possible. In addition, the attention mechanism is employed to discriminate the contributions of ensemble diversity measures with different feature interaction orders. Finally, in the objective function, we introduce both ensemble diversity and prediction accuracy to exploit the merits of ensemble learning.

## 2.3 Ensemble Learning

Ensemble learning, in which a set of base learners is combined, has been proven theoretically and empirically to generate more accurate predictions than a single learner does [41]. However, constructing them is not a trivial task. The key is to construct base learners that are as accurate and diverse as possible.

In recent years, although there is no generally accepted definition of diversity, different ensemble diversity measures have been proved effective and introduced in ensemble learning to make a trade-off between ensemble diversity and prediction accuracy in various areas. For example, Liu et al. [29] introduced a correlation penalty term into the objective function, which measures the diversities of each base learner against the entire ensemble. Zhang et al. [30] proposed to calculate the prediction differences between each pair of base learners, which maximizes the accuracies of base learners on the labeled data and helps augment the diversity among them on the unlabeled data. Liu et al. [31] proposed a multiple kernel k-means clustering algorithm with a matrix-induced regularization, in which the inner products of each pair of kernels are measured to enhance the diversity of selected kernels. Guo et al. [32] introduced a special form of the inner product measurement, termed as exclusivity, which uses the Hadamard product and encourages two base learners to be as orthogonal as possible. Yu et al. [33] proposed an ensemble of support vector machines with an imposed diversity constraint, which calculates the cosine similarity between different base learners in a mathematical programming framework. All these works have justified the effectiveness of constructing different diversity measures in ensemble learning.

## 2.4 Attention Mechanism

Recently, the attention mechanism, which has the capability to focus on the selective parts of the whole perception space as needed, has been widely employed in various areas, including machine translation, computer vision, and recommendation systems. For example, Bahdanau et al. [34] were the first to introduce the attention mechanism into machine translation, whose model outperforms the conventional models significantly. Hou et al. [35] proposed a framework to learn the weight of the object-level regions subject to image aesthetic assessment through the attention mechanism. Li et al. [36] employed a transformer with multi-head self-attention for feature learning, which provides interpretable insights of the results in click-through rate prediction. Yuan et al. [37] proposed an attention-based context-aware sequential recommendation model to distinguish the importance of each item in the rating sequence. Liang et al. [7] modeled feature interactions from different views through the feature-level and view-level attentions.

To exploit the merits of the attention mechanism, we introduce it in the proposed model DexDeepFM, which aims to learn attentive weights for the ensemble diversity measures of each hidden layer with different feature interaction orders.

## 3 METHODOLOGY

This section gives the detailed description of the proposed model DexDeepFM. Figure 1 presents the framework of DexDeepFM. Firstly, an embedding layer is applied on the sparse input data to attain the low dimensional dense embedding vector. Then, the interaction component and the deep component are built, which share the same input embedding layer. In addition, a linear component is applied on the input data. Finally, we combine the above components and send them to the output unit.

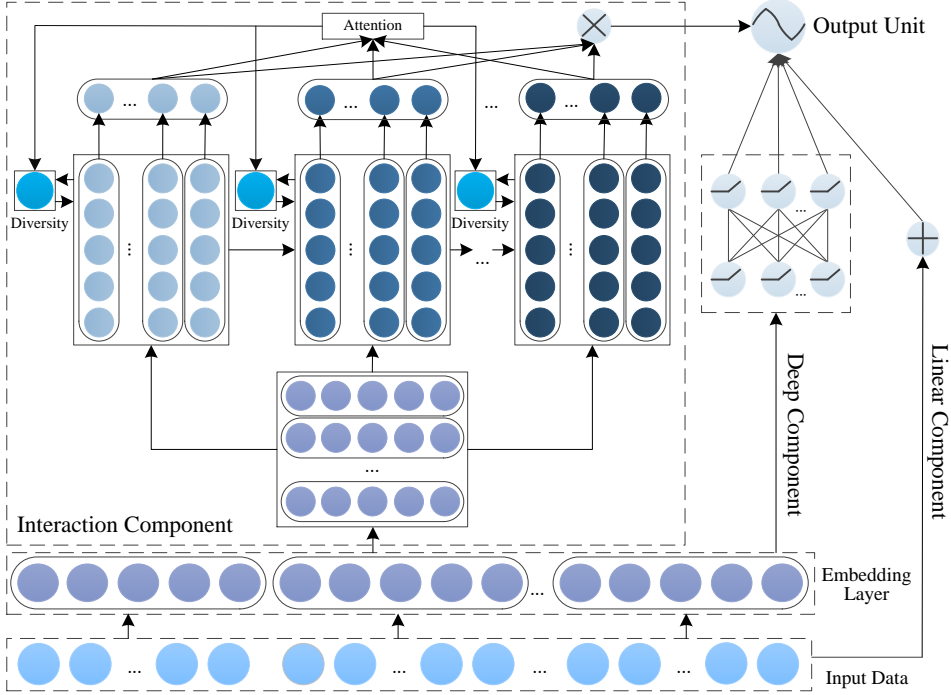

Figure 1: The framework of the proposed model DexDeepFM.

### 3.1 Problem Definition

Before introducing the technical details, we first define some basic notations and the task.

**Notation**: Lower-case letters denote scalars, bold lower-case letters denote vectors, and bold upper-case letters denote matrices. |…| denotes the concatenation operator. $^T$ denotes the transpose operator. ∘ denotes the Hadamard product. $\mathbb{R}^d$ denotes the $d$-dimensional Euclidean space.

**Definition**: Given a dataset $D = \{(x_1, y_1), \ldots, (x_q, y_q)\}$ including $q$ samples, where $x$ is the feature vector of a user, an item, and their respective contexts. $y \in \{0, 1\}$ is the corresponding label of the user response. The task is to build a model, which can predict the positive response probability of a user on a specific item based on particular contextual information. Specifically, $y = 1$ indicates the positive response of the user with regard to a specific item (e.g., purchasing goods or clicking on advertisements), and $y = 0$ indicates the negative response.

### 3.2 Embedding Layer

The structure of the embedding layer is shown in Figure 2. A common practice is to embed the high dimensional and sparse input data into a low dimensional dense real-value vector, which is widely employed in existing works [8-9, 17].

Each feature (e.g., "Gender" and "Film Genre") in the input data is represented as a vector of one-hot or multi-hot encoding. We directly learn the embedding vectors of each feature as the parameters of size $d$. Specifically, for the feature with multi-hot encoding, we sum up the corresponding embedding vectors. For example, in the instance ("Film Genre"="Sci-Fi" and "Adventure"), the embedding vectors of "Sci-Fi" and "Adventure" are summed up as the embedding vector of feature "Film Genre". For the feature with one-hot encoding, the corresponding embedding vector is adopted. For example, in the instance ("Gender"="Female"), the embedding vector of "Female" is used as the embedding vector of feature "Gender". Finally, as the output of the embedding layer, we concatenate all the embedding vectors into one vector $x_0$, which is formulated as Equation 1:

$$x_0 = |x_{\text{embed\_1}}, x_{\text{embed\_2}}, \ldots, x_{\text{embed\_m}}|, \qquad (1)$$

where $x_{\text{embed\_m}} \in \mathbb{R}^d$ is the embedding vector of a feature, $m$ is the number of features, and $d$ is the dimension of embedding vector, e.g., $d$ is set as 5 in Figure 2.

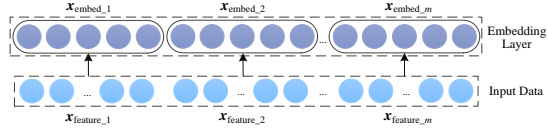

Figure 2: The structure of the embedding layer.

### 3.3 Interaction Component

Based on the interaction component in [17], we present an ensemble diversity enhanced interaction network (DEIN) in DexDeepFM. To achieve feature interactions at the vector-wise level, firstly, matrix $\mathbf{X}^0 \in \mathbb{R}^{m \times d}$ is created by transforming the output of the embedding layer (i.e., the long concatenated vector $x_0$) into a matrix, which is then used as the initial layer for interactions. In $\mathbf{X}^0$, each row denotes an embedding vector of a feature, i.e., the $i$-th row equals to $x_{\text{embed\_i}}$ of $x_0$ in Equation 1.

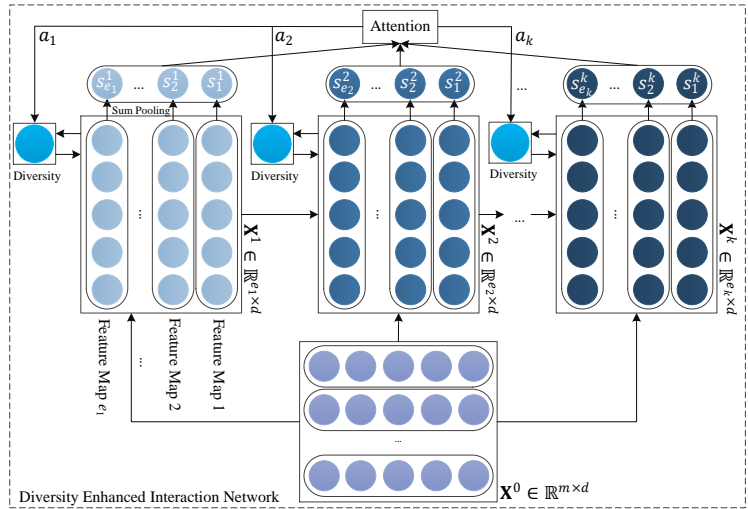

Figure 3: The detailed illustration of interaction component DEIN in DexDeepFM.

As shown in Figure 3, each hidden layer in DEIN is represented as a matrix, and the next hidden layer is generated from the interactions between the last and initial layers, which is formulated as Equation 2:

$$\mathbf{X}_{l,*}^{k+1} = \sum_{i=1}^{e_k} \sum_{j=1}^{m} \mathbf{W}_{i,j}^{k+1,l} \left( \mathbf{X}_{i,*}^{k} \circ \mathbf{X}_{j,*}^{0} \right), \tag{2}$$

where $k \in \mathbb{Z}$ and $1 \le l \le e_{k+1}$. $\mathbf{X}^{k+1} \in \mathbb{R}^{e_{k+1} \times d}$ is the $(k+1)$-th hidden layer. $e_{k+1}$ is the number of embedding vectors in the $(k+1)$-th hidden layer, i.e., there are $e_{k+1}$ different feature maps in $\mathbf{X}^{k+1}$. To make it more clear, the superscript $k$ in $\mathbf{X}_{l,*}^{k}$ means the layer order in DEIN, and the subscript $l$ means the row order in the matrix. In addition, $\mathbf{W}^{k+1,l} \in \mathbb{R}^{e_k \times m}$ denotes the parameters of the $l$-th feature map, and $\circ$ denotes the Hadamard product, i.e., the dimension-wise product.

Specially, $e_0$ equals to $m$, i.e., the number of embedding vectors in $\mathbf{X}^0$, and the first hidden layer $\mathbf{X}^1 \in \mathbb{R}^{e_1 \times d}$ is generated from the interactions between two initial layers $\mathbf{X}^0$, which is formulated as Equation 3:

$$\mathbf{X}_{l,*}^{1} = \sum_{i=1}^{m} \sum_{j=1}^{m} \mathbf{W}_{i,j}^{1,l} \left( \mathbf{X}_{i,*}^{0} \circ \mathbf{X}_{j,*}^{0} \right). \tag{3}$$

With each hidden layer, we add up all elements for each feature map via sum pooling, which is formulated as Equation 4:

$$s_l^k = \sum_{j=1}^{d} \mathbf{X}_{l,j}^{k}, \tag{4}$$

where $1 \le l \le e_k$. As shown in the top of Figure 3, after sum pooling, the output of the $k$-th hidden layer is transformed to a vector $\boldsymbol{s}^k$, which is formulated as Equation 5:

$$\boldsymbol{s}^k = |s_1^k, s_2^k, \ldots, s_{e_k}^k|. \tag{5}$$

Finally, we concatenate the output vectors of all hidden layers into a long vector $\boldsymbol{x}_{\text{dein}}$ and use it as the output of DEIN, which is formulated as Equation 6:

$$\boldsymbol{x}_{\text{dein}} = |\boldsymbol{s}^1, \boldsymbol{s}^2, \ldots, \boldsymbol{s}^K|, \tag{6}$$

where $K$ denotes the number of hidden layers in DEIN, i.e., the depth of the network.

*3.3.1 Ensemble Diversity Measure*

Although it is relatively easy to evaluate the accuracy of a model, there is so far no commonly accepted definition of diversity, and various ensemble diversity measures have been empirically proposed, including Negative Correlation Learning (NCL) [29], which is widely used [38-40]. Instead of creating unbiased individual learners, NCL creates negatively correlated base learners to encourage specialization and cooperation. In addition, it measures the diversities of each base learner against the entire ensemble.

Inspired by NCL, the ensemble diversity measure is designed in each hidden layer of DexDeepFM. We regard different feature maps of a hidden layer in DEIN as the outputs of different base learners by analogy, as shown in Figure 3. For each hidden layer, we expect each feature map to be as diverse as possible. For example, in the first hidden layer $\mathbf{X}^1$, the distances between each feature map $\mathbf{X}_l^1$ ($1 \le l \le e_1$) should be as far as possible. Specifically, the ensemble diversity measure in each hidden layer is formulated as Equations 7-9:

$$Divs(\mathbf{X}^k) = \frac{1}{e_k} \sum_{l=1}^{e_k} dis(\mathbf{X}_{l,*}^{k}, \overline{\boldsymbol{x}^k}), \tag{7}$$

$$dis\left(\mathbf{X}_{l,*}^k, \overline{\boldsymbol{x}^k}\right) = \sqrt{\sum_{j=1}^{d}\left(\mathbf{X}_{l,j}^k - \overline{\boldsymbol{x}_j^k}\right)^2}, \tag{8}$$

$$\overline{\boldsymbol{x}^k} = \frac{1}{e_k}\sum_{l=1}^{e_k}\mathbf{X}_{l,*}^k, \tag{9}$$

where $Divs(\mathbf{X}^k)$ denotes the ensemble diversity measure of the $k$-th hidden layer and $dis(\mathbf{X}_{l,*}^k, \overline{\boldsymbol{x}^k})$ measures the Euclidean distance between $\mathbf{X}_{l,*}^k$ and $\overline{\boldsymbol{x}^k}$. $\overline{\boldsymbol{x}^k} \in \mathbb{R}^d$ is the average result of all feature maps in the $k$-th hidden layer. $e_k$ is the number of feature maps.

### 3.3.2 Attention Mechanism

Since the attention mechanism is helpful in capturing the different importance of different parts in a task, which has shown its effectiveness in various areas (e.g., machine translation [34], computer vision [35], and recommendation systems [37]), we adopt the attention mechanism to learn attentive weights for the ensemble diversity measures. The weights can be interpreted as the different contributions of ensemble diversity measures with different feature interaction orders. With the attention mechanism [43], we rewrite the ensemble diversity measure in Equation 7 as follows:

$$Divs'(\mathbf{X}^k, a_k) = a_k \cdot \frac{1}{e_k}\sum_{l=1}^{e_k} dis(\mathbf{X}_{l,*}^k, \overline{\boldsymbol{x}^k}), \tag{10}$$

where $a_k$ is the attentive weight for the ensemble diversity measure of a hidden layer, which can be explained as the different importance in the task. $Divs'(\mathbf{X}^k, a_k)$ denotes the ensemble diversity measure of the $k$-th hidden layer with the attentive weight. According to [16], we employ a multi-layer perceptron to learn attentive weight $a_k$, which is formulated as Equations 11-12:

$$a_k' = \boldsymbol{h}^{\mathrm{T}}\mathrm{ReLU}(\mathbf{W}\boldsymbol{s}^k + \boldsymbol{b}), \tag{11}$$

$$a_k = \frac{\exp(a_k')}{\sum_{k=1}^{K}\exp(a_k')}, \tag{12}$$

where $\boldsymbol{h} \in \mathbb{R}^a$, $\mathbf{W} \in \mathbb{R}^{a \times e_k}$, and $\boldsymbol{b} \in \mathbb{R}^a$ are the parameters to be learned and shared in each hidden layer. $\boldsymbol{s}^k$ is defined by Equation 5. As shown in the top of Figure 3, the input of the attention network is the sum pooling results of each hidden layer $\boldsymbol{s}^k$. Following [16], we employ rectified linear units (ReLU) as the activation function and use the softmax function to normalize the attentive weights.

### 3.4 Deep Component

The deep component in DexDeepFM, which is employed to learn high-order feature interactions implicitly, is a fully-connected feed-forward neural network. The output of the embedding layer $\boldsymbol{x}_0$ is fed into the neural network, and the forward process of a deep layer is formulated as Equation 13:

$$\boldsymbol{x}_{l+1} = \sigma(\mathbf{W}_l\boldsymbol{x}_l + \boldsymbol{b}_l), \tag{13}$$

where $\boldsymbol{x}_l$, $\mathbf{W}_l$, and $\boldsymbol{b}_l$ are the output, weight, and bias of the $l$-th hidden layer, respectively. $l$ is the layer depth, and $\sigma$ is the activation function.

### 3.5 Output Unit

To learn both low-order and high-order feature interactions explicitly and implicitly, hybrid structures are also exploited in DexDeepFM. As shown in Figure 1, interaction component, deep component, and linear component are combined and fed into the output unit. The result of output unit is formulated as Equation 14:

$$\hat{y} = \sigma(\mathbf{W}_{\text{dein}}^T \boldsymbol{x}_{\text{dein}} + \mathbf{W}_{\text{deep}}^T \boldsymbol{x}_{\text{deep}} + \mathbf{W}_{\text{linear}}^T \boldsymbol{x}_{\text{raw}} + \boldsymbol{b}), \quad (14)$$

where $\sigma$ is the sigmoid function. $\boldsymbol{x}_{\text{dein}}$ and $\boldsymbol{x}_{\text{deep}}$ are the output of interaction component and deep component, respectively. $\boldsymbol{x}_{\text{raw}} = |\boldsymbol{x}_{\text{feature}_1}, \boldsymbol{x}_{\text{feature}_2}, \dots, \boldsymbol{x}_{\text{feature}_m}|$ is the raw features in the input data. $\mathbf{W}_{\text{dein}}$, $\mathbf{W}_{\text{deep}}$, $\mathbf{W}_{\text{linear}}$, and $\boldsymbol{b}$ are both learnable parameters that are trained jointly in the model.

### 3.6 Learning

To estimate model parameters, we need to minimize the following objective function, which considers both prediction accuracy and ensemble diversity:

$$\mathcal{L} = \mathcal{L}_l - \lambda_d \mathcal{L}_d + \lambda_n \|\Theta\|, \quad (15)$$

where $\mathcal{L}_l$ denotes the log loss. $\mathcal{L}_d$ denotes the ensemble diversity loss, and $\lambda_d$ is the parameter that controls the trade-off between $\mathcal{L}_l$ and $\mathcal{L}_d$. To further avoid overfitting, L2 regularization is also considered. $\lambda_n$ is the corresponding regularization term, and $\Theta$ denotes all the parameters in DexDeepFM.

Specifically, log loss $\mathcal{L}_l$, which is widely used in the related works [8-9, 17] for binary classification problems, is formulated as Equation 16:

$$\mathcal{L}_l = -\frac{1}{q} \sum_{(x,y)\in D} \{y \log \hat{y} + (1-y) \log(1-\hat{y})\}, \quad (16)$$

where $D$ is the training set that includes $q$ training samples, as described in Section 3.1. $y$ is the true label indicating the user response, and $\hat{y}$ is the output probability that is calculated by Equation 14.

In addition, with the ensemble diversity measure in each hidden layer of DEIN by Equation 10, we use it as a constraint in the objective function to exploit the merits of ensemble learning. Ensemble diversity loss $\mathcal{L}_d$, which is expected to be maximized, is formulated as Equation 17:

$$\mathcal{L}_d = \frac{1}{q} \sum_{(x,y)\in D} \sum_{k=1}^{K} Divs'(\mathbf{X}^k, a_k). \quad (17)$$

## 4 EXPERIMENTS

In this section, we present the experiments to empirically evaluate the proposed model DexDeepFM. First, three datasets used in the experiments are introduced. Second, experimental settings are given. Finally, experimental results and discussions are presented.

### 4.1 Datasets

We evaluate the proposed model on three public real-world datasets: The Criteo Display Advertising Challenge Dataset[1] (Criteo), MovieLens Dataset[2] (MovieLens), and Avazu Click-Through Rate Prediction Dataset[3] (Avazu).

---

[1] https://labs.criteo.com/2014/02/download-kaggle-display-advertising-challenge-dataset/
[2] https://grouplens.org/datasets/movielens/
[3] http://www.kaggle.com/c/avazu-ctr-prediction

**Criteo**. It is a widely used benchmark dataset for click-through rate prediction, which contains 45 million users' click records on given advertisements. There are 13 integer attributes, which are discretized according to [17], and 26 categorical attributes.

**MovieLens**. It contains over 1 million ratings of approximately 3,900 movies made by 6,040 MovieLens users. Since it is originally designed for explicit rating prediction, we remove samples with a rating of three and binarize it for positive response prediction following [42]. For samples with a rating less than three, we regard them as negative records, and we regard the rest samples as positive ones.

**Avazu**. It contains over 40 million advertisement click records. There are 22 categorical attributes (e.g., app category, device model, and device type) that indicate the elements of a single advertisement impression.

## 4.2 Experimental Settings

The 5-fold cross-validation is used as the evaluation strategy in the experiments. We split all data into five groups randomly, and take four groups of data as the training set (90%) and the validation set (10%), the remaining group as the test set in turn.

We use Logloss, i.e., cross entropy, and AUC, i.e., the area under the ROC curve, as the evaluation metrics in the following experiments to evaluate the performance of the proposed model, which are commonly adopted in existing works [8-9, 17]. Specifically, Logloss measures the performance of a classification model by comparing the predicted probability with the true label, which is formulated as Equation 18:

$$\text{Logloss} = y \log p + (1 - y) \log(1 - p), \tag{18}$$

where $p$ is the output probability of the prediction. While AUC means the probability that a random selection from the positive class will have a score higher than a randomly selected one from the negative class [44], which is formulated as Equation 19:

$$\text{AUC} = \frac{\sum_{i \epsilon D^-} \sum_{j \epsilon D^+} \text{I}(y(i) < y(j))}{|D^-| \cdot |D^+|}, \tag{19}$$

where $\text{I}(z)$ is an indicator function that returns 1 if $z$ is true and returns 0 if $z$ is false. $D^-$ denotes the set of negative samples and $D^+$ denotes the set of positive samples. The higher the AUC is, the better the model is at prediction.

We implement DexDeepFM using TensorFlow, and the code is released on GitHub[4]. In the implementation, we use Adam [45] as the optimizer. Following [17], the dimension of embedding vector $d$ is set as 10, the initial learning rate is set as 0.001, the epoch number is set as 10, and the batch size is set as 4096. For the layer depth in the deep component, we select it among {1, 2, 3, 4} and the optimal setting is 2. For the number of neurons in the deep component, we select it among {100, 200, 300, 400} and the optimal setting is 300. Through the parameter tuning, the number of hidden layers in DEIN is set as 3, the number of feature maps in each hidden layer of DEIN is set as 200, and trade-off parameter $\lambda_d$ is set as 0.7 (detailed in Section 4.3.1).

It is worth noting that, as pointed out in the existing works [42], an improvement of 0.001 in the performance is considered as practically significant for the result in click-through rate prediction. Considering the large user base of real-world scenarios in the industry, a small improvement for the result can lead to a large increase in the revenue. In addition, the pairwise t-test is adopted to analyze the statistical significance of observed differences.

---

[4] https://github.com/hys-451/dexdeepfm

## 4.3 Experimental Results

### 4.3.1 Parameter Tuning

We investigate the influence of important parameters on the performance of the proposed model, including the number of hidden layers, the number of feature maps in each hidden layer, and the parameter that controls the trade-off between prediction accuracy and ensemble diversity.

To explore the impact of the number of hidden layers, we vary it from 1 to 4 in DEIN. The experimental results on the Criteo and MovieLens datasets are shown in Figures 4a and 5a, from which we can find that with the increase of the number of hidden layers in DEIN, the model performance increases first and then decreases. When the number is 3, the model achieves the lowest Logloss and the highest AUC performances in both datasets. Therefore, we set the number of hidden layers in DEIN as 3.

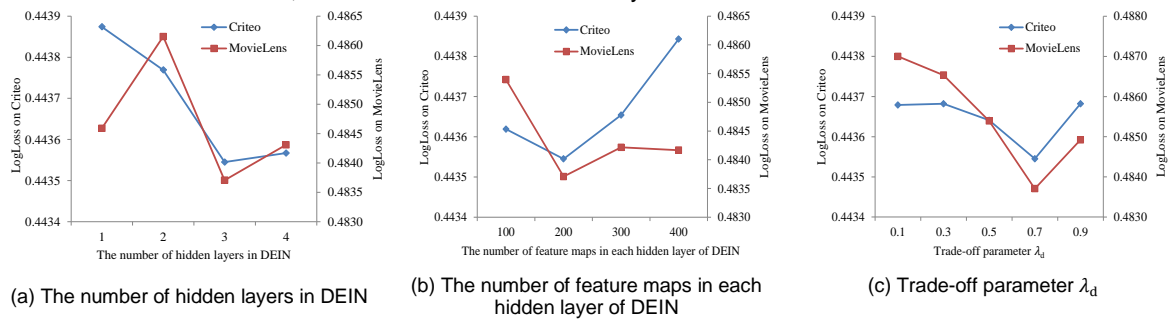

(a) The number of hidden layers in DEIN  (b) The number of feature maps in each hidden layer of DEIN  (c) Trade-off parameter $\lambda_d$

Figure 4: The impact of hyper-parameters on Logloss performance.

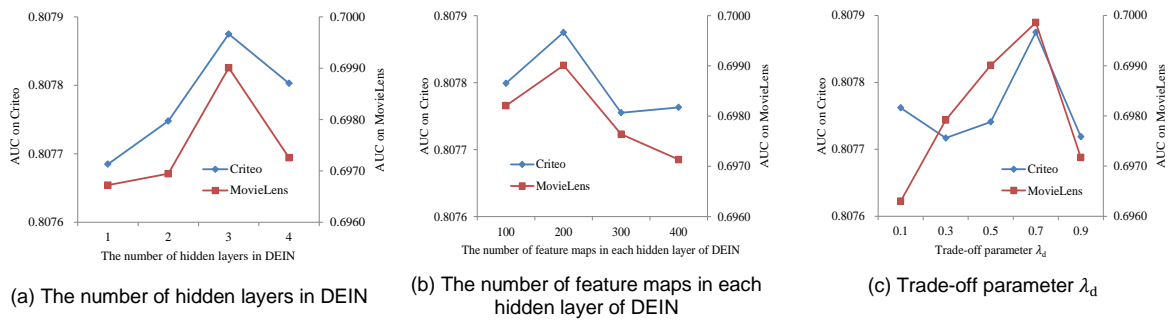

(a) The number of hidden layers in DEIN  (b) The number of feature maps in each hidden layer of DEIN  (c) Trade-off parameter $\lambda_d$

Figure 5: The impact of hyper-parameters on AUC performance.

For the number of feature maps in each hidden layer of DEIN, we vary it from 100 to 400 in the experiments. The experimental results on the Criteo and MovieLens datasets are shown in Figures 4b and 5b, from which we can find similar tendencies. As the number of feature maps in each hidden layer of DEIN increases, the model performance ascends at the beginning and then descends. When the number of feature maps in each hidden layer is 200, the model achieves the best performance. The reason may be that the larger the number of feature maps is, the more parameters are needed to learn, which may cause overfitting. Therefore, we set the number of feature maps in each hidden layer of DEIN as 200.

Parameter $\lambda_d$ that controls the trade-off between prediction accuracy and ensemble diversity also plays an important role in the model performance. We increase the value of $\lambda_d$ from 0.1 to 0.9 in the experiments. The experimental results on the Criteo and MovieLens datasets are shown in Figures 4c and 5c, from which we can find that the model performance improves with the increase of $\lambda_d$. When $\lambda_d$ exceeds 0.7, the model performance decreases. The reason may be that if $\lambda_d$ is set too large, the diversity requirement is dominant over the accuracy requirement. Finally, we set $\lambda_d$=0.7 as a preferable balance point to make a trade-off between accuracy and diversity in the following experiments.

*4.3.2 Variant Comparison*

We design three simplified variants of DexDeepFM to explore the effectiveness of each model component, including DexDeepFM w/o Diversity, DexDeepFM w/o Attention, and DexDeepFM w/o Deep, which remove the ensemble diversity measure, attention mechanism, and deep component, respectively.

The experimental results on both datasets are shown in Table 1, from which we can find that:

1) DexDeepFM w/o Diversity, which is the same as xDeepFM, performs the worst, which indicates the effectiveness of considering ensemble diversity and prediction accuracy jointly in the objective function.

2) The performance of DexDeepFM w/o Attention is the closest to that of DexDeepFM. The reason may be that the number of hidden layers is just 3, which limits the effects of the attention mechanism.

3) The performance of DexDeepFM w/o Deep drops, which justifies that the deep component in the model still plays an important role by learning high-order feature interactions.

Table 1: The performances of the simplified variants of DexDeepFM

| Variant | Criteo | | MovieLens | | Avazu | |
|---|---|---|---|---|---|---|
| | Logloss | AUC | Logloss | AUC | Logloss | AUC |
| DexDeepFM w/o Diversity | 0.4457* | 0.8057* | 0.4891* | 0.6895* | 0.3861* | 0.7718* |
| DexDeepFM w/o Attention | 0.4438* | 0.8078 | 0.4841* | 0.6979* | 0.3834* | 0.7749* |
| DexDeepFM w/o Deep | 0.4450* | 0.8064* | 0.4847* | 0.6984* | 0.3844* | 0.7746* |
| DexDeepFM | **0.4435** | **0.8079** | **0.4837** | **0.6990** | **0.3832** | **0.7751** |

\* denotes DexDeepFM is statistically superior to the compared variant (pairwise t-test at a 95% significance level).

*4.3.3 Baseline Comparison*

To evaluate the performance of DexDeepFM, we conduct an experiment to compare it with other baseline models. The compared models are as follows:

FM [11]: FM exploits the idea of matrix factorization and models feature interactions as the inner products of latent vectors by projecting features into a low dimension latent space.

FNN [14]: FNN introduces a pre-trained FM and applies DNNs to capture high-order feature interactions.

AFM [16]: AFM introduces the attention mechanism to distinguish the different importance of feature interactions based on FM.

NFM [27]: NFM introduces a Bi-Interaction pooling operation layer to deepen FM under the neural network framework for capturing high-order feature interactions.

Wide & Deep [8]: Wide & Deep combines the linear model (the wide part) and DNNs (the deep part) jointly, which captures both low-order and high-order feature interactions together.

IFM [47]: IFM enhances FM by explicitly considering the impact of each individual input on the representations of features and learns a unique input-aware factor for the same feature in different instances.

DeepFM [9]: DeepFM introduces a FM layer to replace the wide part in [8] and employs DNNs to build hybrid structures that exploit the merits of low-order and high-order feature interactions.

xDeepFM [17]: xDeepFM introduces an interaction network to replace the FM layer in DeepFM, which can learn explicit high-order feature interactions at the vector-wise level.

The experimental results on both datasets are shown in Table 2, from which we can find that:

1) DexDeepFM achieves the best performance in all comparative methods. Specifically, it outperforms traditional feature interaction models (i.e., FM and AFM) dramatically. It also excels the deep learning based models distinctly, including both the single structure (i.e., FNN, NFM, and IFM) and hybrid structures (i.e., Wide & Deep and DeepFM) based models.

2) With respect to the second-best model xDeepFM, DexDeepFM significantly outperforms it, which justifies the advantages of introducing the ensemble diversity measure and the attention mechanism in the proposed model.

3) Traditional feature interaction models (i.e., FM and AFM) obtain worse results than deep learning based models. The main reason is that they only exploit the low-order feature interactions, which limits their representation power.

4) For deep learning based models (i.e., FNN, NFM, IFM, Wide & Deep, and DeepFM), FNN and NFM obtain worse performance than others, which is because they focus on learning high-order feature interactions in a single structure and ignore the low-order feature interactions that are also important for the results. In addition, the results of IFM give us insights that a single structure with diverse and accurate representations of features can also achieve competitive performance.

Table 2: The performances of compared models

| Model | Criteo | | MovieLens | | Avazu | |
| --- | --- | --- | --- | --- | --- | --- |
| | Logloss | AUC | Logloss | AUC | Logloss | AUC |
| FM | 0.4595 | 0.7886 | 0.5085 | 0.6673 | 0.3942 | 0.7520 |
| FNN | 0.4546 | 0.7937 | 0.4998 | 0.6702 | 0.3904 | 0.7589 |
| AFM | 0.4533 | 0.7948 | 0.4945 | 0.6813 | 0.3912 | 0.7626 |
| NFM | 0.4525 | 0.7980 | 0.4965 | 0.6743 | 0.3898 | 0.7658 |
| Wide & Deep | 0.4522 | 0.7972 | 0.4933 | 0.6807 | 0.3875 | 0.7673 |
| IFM | 0.4484 | 0.7994 | 0.4904 | 0.6852 | 0.3863 | 0.7678 |
| DeepFM | 0.4491 | 0.8023 | 0.4922 | 0.6833 | 0.3880 | 0.7676 |
| xDeepFM | 0.4457 | 0.8057 | 0.4891 | 0.6895 | 0.3861 | 0.7718 |
| DexDeepFM | **0.4435**\* | **0.8079**\* | **0.4837**\* | **0.6990**\* | **0.3832**\* | **0.7751**\* |

\* denotes DexDeepFM is statistically superior to the best baseline (pairwise t-test at a 95% significance level).

*4.3.4 Case Study*

To intuitively reveal the effectiveness of ensemble diversity measures in DexDeepFM, we perform a case study. We use the t-SNE method [46] to visualize the learned feature maps of the first and second hidden layers in the interaction networks of DexDeepFM and xDeepFM.

The visualization results on the MovieLens dataset are shown in Figure 6. We can find that the learned feature maps in xDeepFM, which are represented by the blue cross, are distributed concentratedly. Similar focused distributions can be found in both hidden layers. In contrast, the learned feature maps in the proposed model DexDeepFM, which are represented by the red circle, are distributed much more widely and diversely. It indicates that by introducing the ensemble diversity measure in the interaction network, the model can learn better diverse representations and achieve good ensembles.

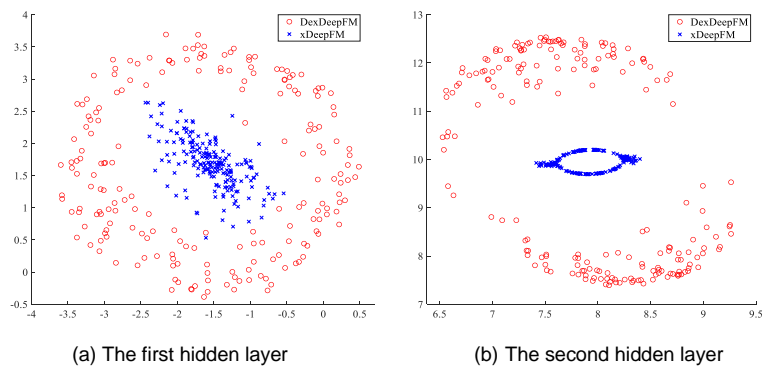

(a) The first hidden layer  (b) The second hidden layer

Figure 6: The t-SNE visualization results on the feature maps of the first and second hidden layers in the interaction networks of DexDeepFM (represented by the red circle) and xDeepFM (represented by the blue cross).

## 5 CONCLUSIONS AND FUTURE WORK

In this work, we propose DexDeepFM, an ensemble diversity enhanced extreme deep factorization machine model, which considers both ensemble diversity and prediction accuracy in the objective function. The ensemble diversity measure is designed in each hidden layer, which ensures that different feature maps can be represented as diverse as possible. In addition, we introduce the attention mechanism to discriminate the importance of ensemble diversity measures with different feature interaction orders. We evaluate the proposed model on three public real-world datasets and the extensive experimental results show the effectiveness of DexDeepFM.

In the future, we will extend this work in the following directions: First, we will design more effective ensemble diversity measures in the interaction network, e.g., designing customized diversity measures for different orders of feature interactions. Second, we will explore the effectiveness of combining two or more ensemble diversity measures in the objective function. Third, we will design other feature interaction network, e.g., a hierarchical interaction network, to control the range of feature interactions.